\documentclass{article}

\usepackage[preprint]{neurips_2025}

\usepackage[utf8]{inputenc} 
\usepackage[T1]{fontenc}    
\usepackage{hyperref}       
\usepackage{url}            
\usepackage{booktabs}       
\usepackage{amsfonts}       
\usepackage{nicefrac}       
\usepackage{microtype}      
\usepackage{xcolor}         

\usepackage{subcaption}
\usepackage{graphicx}
\usepackage{multirow}
\usepackage{wrapfig}

\usepackage{amsmath,amsfonts,bm}

\def\eqref#1{equation~\ref{#1}}

\def\1{\bm{1}}

\DeclareMathAlphabet{\mathsfit}{\encodingdefault}{\sfdefault}{m}{sl}
\SetMathAlphabet{\mathsfit}{bold}{\encodingdefault}{\sfdefault}{bx}{n}

\newcommand{\Figure}[1]{Fig.~\ref{#1}}
\newcommand{\Table}[1]{Table~\ref{#1}}
\newcommand{\Section}[1]{Sec.~\ref{#1}}

\newcommand{\bs}{\boldsymbol}

\title{Grazing Detection using Deep Learning \\ and Sentinel-2 Time Series Data}

\author{%
  Aleksis Pirinen  \thanks{RISE Research Institutes of Sweden $^\dagger$Climate AI Nordics $^\ddagger$Swedish Centre for Impacts of Climate Extremes}$ \ \ \  ^\dagger  \  ^\ddagger$  \\
  \texttt{aleksis.pirinen@ri.se} 
  \And
  Delia Fano Yela $^*$ $^\dagger$  \\
   \texttt{
delia.fano.yela@ri.se} 
   \AND
   Smita Chakraborty $^*$  $^\dagger$   \\
   \texttt{
smita.chakraborty@ri.se} 
   \And
   Erik Källman $^*$ \\
   \texttt{
erik.kallman@ri.se} \\
}

\begin{document}

\maketitle

\begin{abstract}
Grazing shapes both agricultural production and biodiversity, yet scalable monitoring of where grazing occurs remains limited. We study seasonal grazing detection from Sentinel-2 L2A time series: for each polygon-defined field boundary, April–October imagery is used for binary prediction (grazed / not grazed). We train an ensemble of CNN-LSTM models on multi-temporal reflectance features, and achieve an average F1 score of 77\% across five validation splits, with 90\% recall on grazed pastures. Operationally, if inspectors can visit at most 4\% of sites annually, prioritising fields predicted by our model as \emph{not grazed} yields about \textbf{5$\times$} more confirmed non-grazing sites than random inspection. These results indicate that coarse-resolution, freely available satellite data can reliably steer inspection resources for conservation-aligned land-use compliance. Code and models are publicly available at \url{https://github.com/aleksispi/pib-ml-grazing}.
\end{abstract}

\section{Introduction}\label{sec:intro}
%
%
%
%
%
Grazing is central to sustainable agriculture and biodiversity, yet verifying where grazing occurs remains costly and scales poorly when based on field inspections or self-reporting. Many countries, e.g.~EU states under upcoming nature restoration laws, are in need for reliable, large-scale assessments to support compliance, efficient land use, and ecological stewardship, which motivates the need for automated, data-driven monitoring. In this work -- conducted as an applied project jointly with the Swedish Board of Agriculture (SBA), Sweden's authority for overseeing, among other things, grazing activity in Swedish pastures -- we study seasonal grazing detection using Sentinel-2 L2A time series combined with machine learning (ML). Sentinel-2 provides multi-spectral, frequent-revisit imagery,
which enables vegetation dynamics to reveal whether pastures were grazed during a season. We frame the task as time series classification at the field-polygon level.

Our work fits in within recent and contemporary literature such as \cite{vahidi2023estimation, ye2025spatiotemporal, parsons2024machine, correa2024accounting}. However, to the best of our knowledge, ours is the first attempt at leveraging ML for recognizing grazing activity from freely available and coarse-resolution satellite data. Our experimental results indicate that ML-based remote sensing models can vastly improve the efficiency of field inspections of grazing activity, by offering a scalable, cost-effective alternative to manual verification, which in turn can improve resource allocation and decision-making for land-use planning.

\section{Dataset}\label{sec:data}
\begin{figure}[t]
    \centering
    \includegraphics[width=0.99\linewidth]{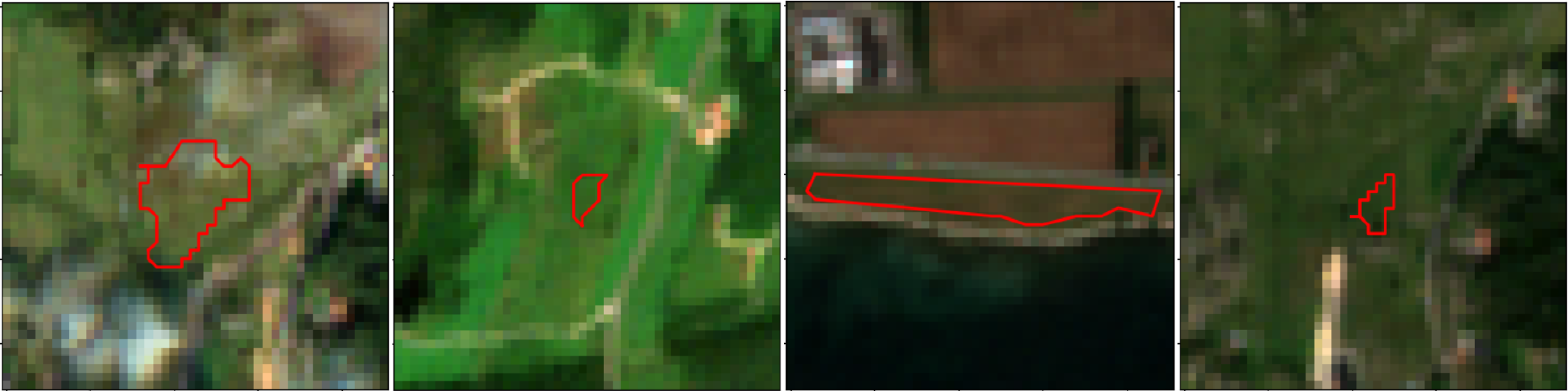}
    \caption{Example RGB-parts of Sentinel-2 L2A time series and field boundaries (polygons).}
    \label{fig:poly-examples}
    \vspace{-6pt}
\end{figure}

Labels and polygons were obtained from the Swedish Board of Agriculture (SBA), for the years 2022 and 2024.
Centered at each polygon, square-shaped (0.45 x 0.45 km) time series Sentinel-2 L2A data was downloaded between April 1st and October 31st for 2022 and 2024, respectively, from the \emph{Digital Earth Sweden (DES)} platform\footnote{https://digitalearth.se} -- see examples in \Figure{fig:poly-examples}. Each time series consists of $T$ images of size $H \times W \times C$, with $H=W=45$ and $C=13$ (all bands of S2-L2A are used).

\subsection{Data preprocessing}\label{sec:data-preproc}
\begin{wrapfigure}[22]{r}{0.35\linewidth}
    \centering
\includegraphics[width=0.80\linewidth]{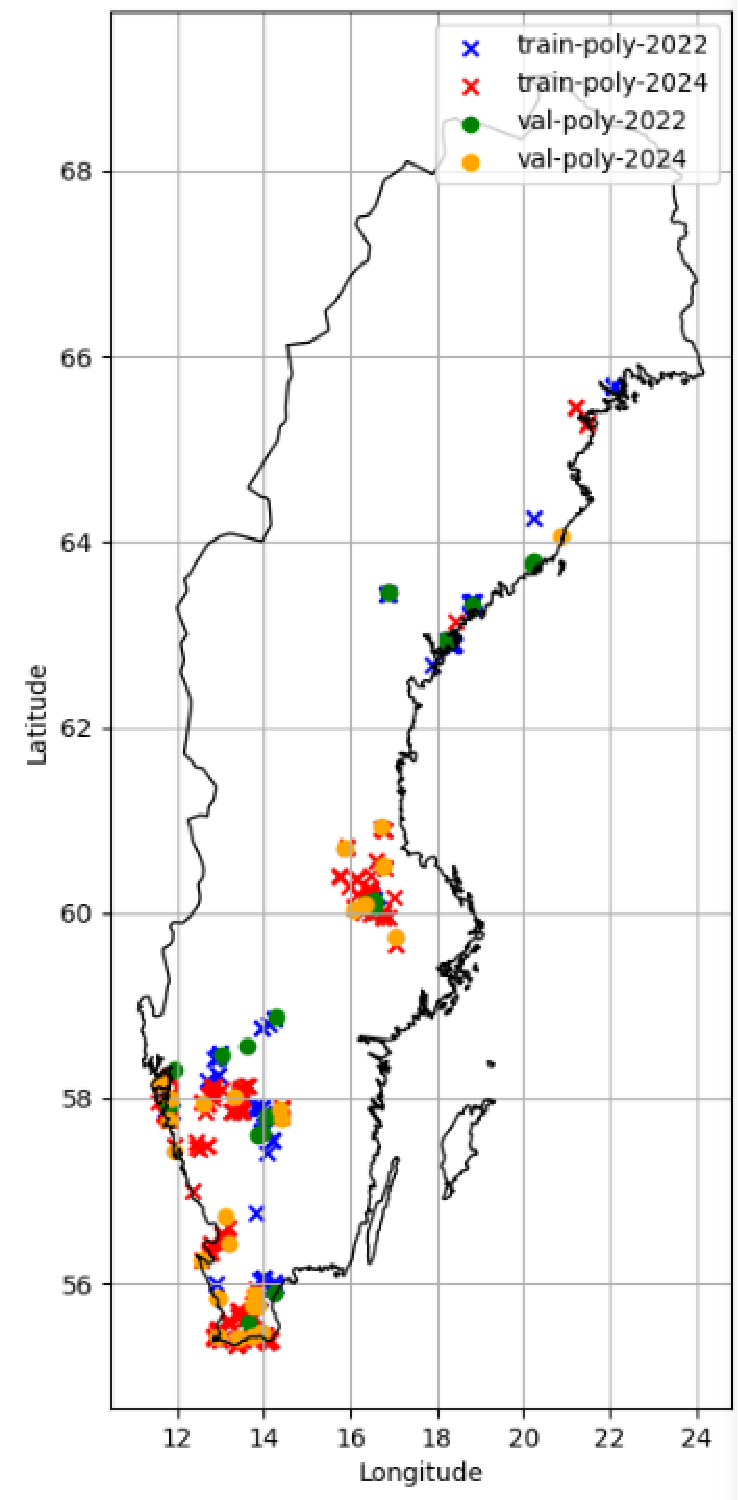}
    \caption{Training and validation polygons across Sweden.}
    \label{fig:sweden-polys}
\end{wrapfigure}
\textbf{Selecting binary labels.} The original 2022 data has labels \emph{Grazing (uncertain)}, \emph{Harvest activity}, \emph{Grazing} and \emph{No activity}. The 2024 data has labels \emph{Lightly grazed}, \emph{Grazed} and \emph{No activity}. In this initial work we focus on the clear case where \emph{grazing} should be differentiated from \emph{no activity}, and pick only polygons with any of these two labels.

\textbf{Removing cloudy images in time series.} We use the method \cite{pirinen2024creating} to predict cloudy pixels in each image. We then remove all images where the polygon contained at least $1\%$ cloudy pixels.

\textbf{Ignoring tiny polygons.} Some polygons are so small that it is not reasonable to assess if grazing has occurred. We therefore discard polygons smaller than $3\times3$ pixels ($30\times 30$ meters).

\subsection{Machine learning-ready dataset}\label{sec:data-ml-ready}
The final ML-ready dataset looks as follows: \textbf{(i)} 108 polygons for 2022, 57 labeled  \emph{grazing}, 51 labeled \emph{no activity}; \textbf{(ii)} 299 polygons for 2024, 196 labeled \emph{grazing}, 103 labeled \emph{no activity}; \textbf{(iii)} 407 polygons in total, 253 labeled \emph{grazing}, 154 labeled \emph{no activity}.
We first partition the data (407 polygons) into a training and validation set (80\% and 20\% of the data, respectively) -- due to the small dataset size obtained for this work, a separate test set is not created. In lack of a dedicated test set, we resort to cross-validation (see \Section{sec:results}). The first train-val split looks as follows: \textbf{(i)} of the 347 training data points, 223 are labeled \emph{grazing} and 124 as \emph{no activity}; \textbf{(ii)} of the 59 validation data points, 30 are labeled \emph{grazing} and 29 as \emph{no activity}. See \Figure{fig:sweden-polys} for the distribution of these polygons in Sweden (we have ensured that there is no spatial overlap between training and validation polygons, although some appear very close based on this zoomed-out view).

\section{Machine learning approach }\label{sec:method}
Our ML-based grazing classification pipeline works as follows. First, we mask out everything outside the polygon in each image time series, so that the model
focuses on the interior of the polygon (as is found to be beneficial -- see \Section{sec:results}). Next, the data is per-channel normalised to mean 0 and standard deviation 1. Finally, the resulting time series is sent to the ML model to predict \emph{grazing} or \emph{no activity}. This ML model consists of three core modules that are run in the following order: \textbf{(i)} spatial processing of images in the time series using a convolutional block; \textbf{(ii)} temporal processing of image feature maps using a bidirectional LSTM \cite{hochreiter1997long}; and \textbf{(iii)} binary classification based on temporal aggregate from step (ii). We next describe the details of each step.

\textbf{(i) Spatial processing.} In this step, each image in the time series is \emph{independently} fed through a convolutional block, to capture spatial features. This block is a single convolutional layer ($7 \times 7$ kernel) with a ReLU activation followed by max-pooling.
The feature maps from the spatial processing are then reshaped (vectorized) to match what is expected in the temporal processing step, described next.

\textbf{(ii) Temporal processing.} Given spatial features from the previous step, here a bidirectional LSTM (hidden dimension $d=16$) is used to aggregate information about the time series over time.

\textbf{(iii) Binary classification.} Finally, the final hidden state $\bs{h}_t$ from the temporal processing step above is fed to a fully connected layer, followed by a sigmoid, which results in the predicted probability of grazing.
When the model is deployed (see \Section{sec:training}), a small modification is however used to improve prediction results slightly.
Specifically, the binary classifier instead looks at the last four hidden states $\bs{h}_{t-3}, \dots, \bs{h}_t$, and for each such hidden state an independent binary prediction is obtained. The final prediction is then given by the majority vote of these predictions.

\subsection{Model training and inference}\label{sec:training}
The model is trained using a standard cross-entropy loss for 300 epochs, with a batch size of $10$. We use Adam \cite{kingma2014adam} with default settings and learning rate \verb|3e-4|. Training a single model takes about 45 minutes on an NVIDIA GeForce RTX 3090 GPU.
In addition to standard data augmentation (left-right and top-down flipping; random cropping), we found it beneficial (see \Section{sec:results}) to apply temporal dropout on the image time series. More specifically, in each batch we remove random time steps, which increases the variability in time series lengths and time gaps that the model is exposed to (note that time gaps also occur due to the cloud removal described in \Section{sec:data-preproc}). We apply temporal dropout at 50\% random on the time series, with a $35\%$ chance of individual time steps dropping out.

As empirically shown in \Section{sec:results}, we found it beneficial to leverage ensembles of trained models during inference. An ensemble consists of 10 identical model architectures trained from different random initial parameter sets. From the 10 independent binary predictions, a majority vote is used to obtain a final prediction (\emph{grazing} or \emph{no activity}). Also, recall that during inference, the aggregation is performed not only across the 10 individual model predictions, but also for the time step predictions associated with the last four hidden states of the bi-LSTM.
The 10-ensemble runtime is about 3.5 to 6 ms per time series, depending mainly on the length of the time series.

\begin{table}[t]
\centering
\caption{Five-fold cross-validation results (each with 80\% for train and 20\% for val) for our proposed approach. Here, \emph{gz} refers to \emph{grazing} and \emph{no} refers to \emph{no activity}. The approach used is an ensemble of 10 ML models followed by majority voting. The last 3 rows are ablations on split \#2.}\label{t:main-results}
\vspace{5pt}
\scalebox{0.93}{
\begin{tabular}{|c||c|c|c|c|c|c|c|c|}
\hline
\textbf{Train-val split} & \textbf{Acc} & \textbf{F1} & \textbf{Prec} & \textbf{Rec}  & \textbf{Prec-gz} & \textbf{Prec-no} & \textbf{Rec-gz} & \textbf{Rec-no} \\ \hline \hline
\textbf{Split \#1} & 0.797 & 0.794 & 0.810 & 0.795 & 0.750 & 0.870 & 0.900 & 0.690 \\ \hline
\textbf{Split \#2} & 0.770 & 0.765 & 0.791 & 0.768 & 0.718 & 0.864 & 0.903 & 0.633 \\ \hline
\textbf{Split \#3}  & 0.772 & 0.771 & 0.780 & 0.773 & 0.727 & 0.833 & 0.857 & 0.690 \\ \hline
\textbf{Split \#4} & 0.733 & 0.729 & 0.751 & 0.733 & 0.684 & 0.818 & 0.867 & 0.600 \\ \hline
\textbf{Split \#5} & 0.807 & 0.801 & 0.817 & 0.798 & 0.778 & 0.857 & 0.903 & 0.692 \\ \hline
\textbf{Mean} & 0.776 & 0.772 & 0.790 & 0.774 & 0.731 & 0.848 & 0.886 & 0.661 \\ \hline
\textbf{Median} & 0.772 & 0.771 & 0.791 & 0.773 & 0.727 & 0.857 & 0.900 & 0.690 \\ \hline
\hline
\textbf{Single-model} & 0.721 & 0.717 & 0.733 & 0.720 & 0.690 & 0.775 & 0.823 & 0.617 \\ \hline
\textbf{Poly-input} & 0.672 & 0.670 & 0.675 & 0.671 & 0.657 & 0.692 & 0.742 & 0.600 \\ \hline
\textbf{No-temp-aug} & 0.738 & 0.732 & 0.755 & 0.735 & 0.692 & 0.818 & 0.871 & 0.600 \\ \hline
\end{tabular}}
\vspace{-6pt}
\end{table}

\section{Experimental results}\label{sec:results}
Our main results, using a 10-ensemble of ML models as described in \Section{sec:training}, are based on cross-validation over five random train-val splits -- see \Table{t:main-results}. We note that there is some variation in results between splits (e.g.~split \#4 vs \#5). However, the median results suggest that one can expect about $77\%$ F1-score, $79\%$ precision and $77\%$ recall at previously unseen sites. We note that our ML-approach is most reliable at grazing sites, where it obtains a recall \emph{Rec-gz} of $90\%$ (few false negatives). It is however not as reliable at non-grazing sites, with a recall \emph{Rec-no} of $69\%$. However, as shown in \Section{sec:saved-visits}, the practical implication of these results is notable.

\Table{t:main-results} also contains results on split \#2 for three alternative ML approaches: \textbf{(i)} \emph{Single-model} (average across 10 independent runs of the models in the 10-ensemble); \textbf{(ii)} \emph{Poly-input} (main 10-ensemble approach but where we do \emph{not} mask out the context surrounding the polygons -- the models however obtain the polygon boundaries as inputs, to know that is in-field and out-of-field); and \textbf{(iii)} \emph{No-temp-aug} (main 10-ensemble approach but without temporal dropout). The results suggest that \textbf{(i)} ensembling outperforms single-model-inference; \textbf{(ii)} masking out imagery outside polygons is highly beneficial; and \textbf{(iii)} temporal dropout yields better results. Refer to the \textbf{appendix} for further results.

\subsection{In practice: Application on grazing inspections}\label{sec:saved-visits}
\begin{wrapfigure}{r}{0.42\linewidth}
    \centering    \includegraphics[width=0.92\linewidth]{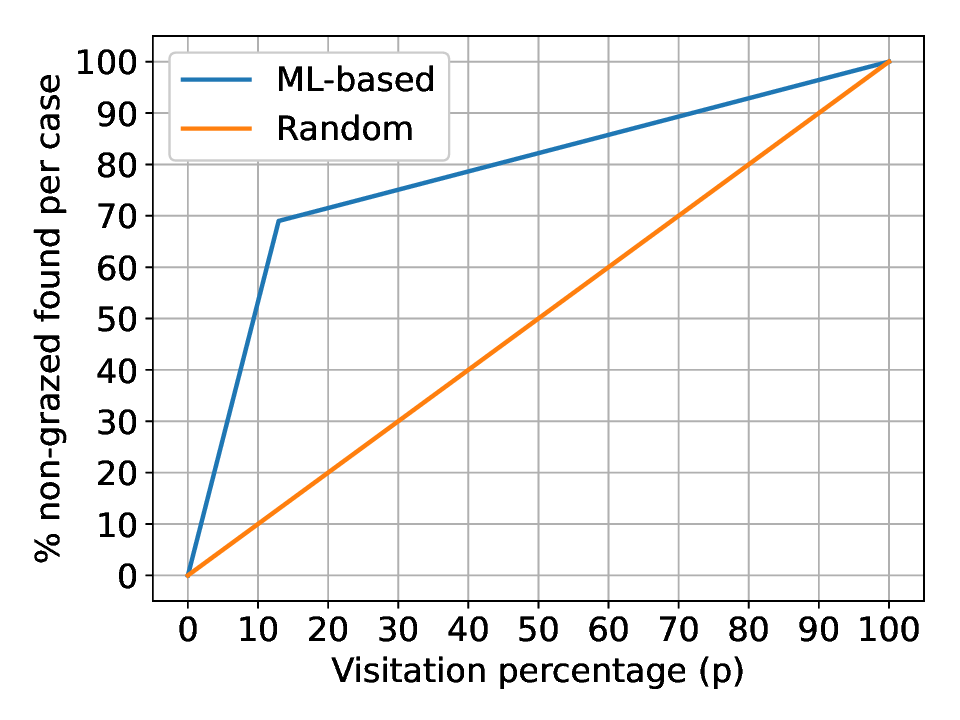}
    \vspace{-6pt}
    \caption{Expected percentage of non-grazing sites found under different visitation percentages $p$, when selecting sites at random either (i) from all sites (orange), or (ii) from the sites predicted as \emph{non-grazed} by our model (blue). We here assume that roughly $5\%$ of all sites are non-grazed. If the visitation percentage $p \leq 13\%$, then our ML-based approach (ii) finds \textbf{5.3x more non-grazing sites}, on average.}
    \vspace{-8pt}
    \label{fig:perc-non-grazed}
\end{wrapfigure}

In Sweden, the Swedish Board of Agriculture (SBA) gives incentives for grazing as it promotes conservation and restoration of pastures. To monitor whether grazing has occurred, on-site inspections are carried out\footnote{The SBA's current approach combines risk-based modeling and random site selection; for the sake of the analysis in this subsection, we simplify and assume random site selection. We compensate for this by significantly overestimating the amount of non-grazed sites, which reduces the relative advantage of our approach.} by domain experts. However, given time and budget constraints, the number of visits the SBA can conduct in a year is very limited. The SBA is mainly interested in discovering sites that have \emph{not} been grazed, as these are the areas for which action should be taken to improve biodiversity (via grazing). Fortunately, such non-grazed sites are quite rare in practice -- it is expected that significantly less that 5\% of all sites per year are \emph{not} grazed -- but this also means that random site selection leads to very few non-grazed sites being discovered. Instead of randomly sampling from all sites in Sweden, we propose to sample from the sites marked as \emph{non-grazed} by our ML approach (until exhausted, then randomly sample from the remaining sites), for which the recall is $69\%$ for \emph{non-grazed} and $90\%$ for \emph{grazed} sites (medians in \Table{t:main-results}).

To concretize the notable improvement of our approach, we present a realistic example. Let us assume that in Sweden there are $10,000$ sites that have claimed grazing incentives, of which only $500$ (5\%) have not been grazed. Our approach marks $345$ ($69\%$, i.e.~the recall for \emph{non-grazed}) of the non-grazed sites as \emph{non-grazed}, but also $950$ ($10\%$, i.e.~one minus the recall for \emph{grazed}) of the $9{,}500$ grazed sites. If the SBA could afford to do on-site inspections of all $1{,}295$ sites marked as \emph{non-grazed} by our approach, they would thus uncover $345$ of the non-grazed sites, compared to the $65$ ($13\%$) uncovered by the same amount of random visits. If only $100$ sites can be visited, by randomly choosing them from the ones identified as non-grazed by the model, about $27$ of those would be truly non-grazed (since $345$ of the $1{,}295$ marked sites, i.e.~$27\%$, are truly non-grazed), uncovering 5.3\% of all non-grazed sites, compared to the 1\% of the current method.
It is thus evident that the presented approach, which relies on the predictions of our ML approach, has a paramount impact in practice, by making the most out of the reduced on-site inspections that the SBA can afford. \Figure{fig:perc-non-grazed} shows the significant improvement our approach can have in the detection of non-grazed sites.

\section{Conclusions}\label{sec:conclusions}
We have shown that seasonal grazing can be detected at field level from Sentinel\mbox{-}2 time series using a CNN--LSTM pipeline. Across five splits, the model attains an average F1 score of 77\% and 90\% recall on grazed fields. Operationally, if inspectors can visit up to at most 4\% of sites per year, targeting fields predicted as \emph{not grazed} yields about \textbf{5$\times$} more confirmed non-grazing sites than random selection, indicating substantial efficiency gains for monitoring and policy enforcement. Future work will focus on: \textbf{(i)} enlarging and diversifying training data across regions and years; \textbf{(ii)} establishing a held-out test set for robust generalization estimates; and \textbf{(iii)} leveraging self-supervised or foundation-model pretraining \cite{brown2025alphaearth, nedungadi2024mmearth, jakubik2025terramind,  blumenstiel2025terramesh, francis2024major, feng2025tessera, astruc2025anysat} -- e.g.~\cite{feng2025tessera} is a recent state-of-the-art foundation model targeted towards agriculture -- to reduce the needs for labeled data and improve transferability.

\subsubsection*{Acknowledgments}
This work was funded by the Swedish National Space Agency (project number 2023-00332). We are also grateful for the support and data provided by the Swedish Board of Agriculture, and Niklas Boke Olén in particular.

\bibliographystyle{unsrt}
\bibliography{bibfile}

\section*{Appendix}
In this appendix we provide additional experimental results on validation split \#1 and \#2 -- see \Table{t:ablations}. If nothing else is specified, the results are always for 10-model ensembles, as in the main paper. More specifically, these approaches are compared in \Table{t:ablations}:
\begin{itemize}
    \item \emph{Main} is the main approach, whose results on multiple train-val splits are also given in \Table{t:main-results}.
    \item \emph{Single-model} is the same as \emph{Main}, but here we look at single-model prediction (no ensemble), where the result reported is the average result one gets by \emph{individually} looking at the results one gets using a single model (average over 10 such results).
    \item \emph{Only-last} is the same as \emph{Main}, except that it only uses the very final time step hidden state as input to the binary classifier (recall that \emph{Main} looks at the median prediction given from the last four time step hidden states instead).
    \item \emph{No-poly} is the same as \emph{Main}, except no information about the polygon is provided (recall that everything outside polygons are masked out in \emph{Main}).
    \item \emph{Poly-input} is the same as \emph{Main}, except instead of masking out image content outside the polygons, the full image content is provided as input, and the polygon geometry is itself provided as an \emph{additional} input.
    \item \emph{No-temp-aug} is the same as \emph{Main}, except no temporal dropout is used during training data augmentation. 
    \item \emph{No-RGB} is the same as \emph{Main}, except it omits the RGB color bands (B02-B04) from the model input (uses 9 instead of 13 channels).
    \item \emph{No-RGB-no-veg} is the same as \emph{Main}, except it omits the RGB color bands (B02-B04) and the vegetation red edge bands (B05-B07) from the model input (uses 6 instead of 13 channels).
    \item \emph{Only-RGB+veg} is the same as \emph{Main}, except it only uses the RGB color bands (B02-B04) and the vegetation red edge bands (B05-B07) as model input (uses 6 instead of 13 channels).
\end{itemize}
The main findings from \Table{t:ablations} are:
\begin{itemize}
    \item Model ensembling yields better results compared to single-model results (e.g.~$0.024$ and $0.048$ increase in F1-score\footnote{Also, have a look at e.g.~the \emph{Rec-gz} metric, i.e.~the recall of true grazing examples. It is higher for the ensemble (\emph{Main})}); see \emph{Main} vs \emph{Single-model}. 
    \item Using only the very last hidden state in the binary classification step reduces task performance compared to aggregating from the last 4 time steps (e.g.~$0.018$ loss in F1-score on both splits); see \emph{Only-last} vs \emph{Main}.
    \item Leveraging information of the polygons is crucial, as omitting all polygon information leads to much worse results (\emph{No-poly} obtains F1-score reductions of $0.056$ and $0.097$ relative to \emph{Main}). Furthermore, comparing \emph{Poly-input} to \emph{Main}, we see that results are about the same for split \#1, but significantly worse on split \#2 (F1-score reduction of $0.095$), and thus worse overall on average. This suggests that the model benefits from masking out the "background information" which is outside the polygo (as is done for \emph{Main}).
    \item Time step dropout as data augmentation has no effect for split \#1, but omitting it for split \#2 leads to somewhat worse results (F1-score reduction of $0.033$); see \emph{Main} vs \emph{No-temp-aug}. It is overall a bit unclear whether time step dropout is actually needed.
    \item Overall, using \emph{all} the Sentinel-2 L2A bands appear to be best, even though omitting the RGB bands seems to have quite little effect on performance (see \emph{Main} vs \emph{No-RGB}; there is only a bit of a drop -- an F1-score reduction of $0.036$ -- in the results on split \#2). Omitting both RGB and the red vegetation edge bands has a stronger negative effect on split \#2 -- an F1-score reduction of $0.085$ -- but it again has no impact on split \#1; see \emph{Main} vs \emph{No-RGB-no-veg}. The worst results are clearly obtained in the setting when only using the RGB and red vegetation edge bands (\emph{Only-RGB+veg}, which yields F1-score reductions of $0.138$ and $0.177$.
\end{itemize}

\begin{table}[t]
\centering
\caption{Various ablation results on split \#1 and \#2 for various ML-based grazing classification approaches explored in this project. Here, \emph{gz} refers to 'grazing' and \emph{no} refers to 'no activity' (so e.g.~\emph{Rec-gz} refers to the average recall across time series with actual grazing in them). If nothing else is specified, each model refers to an ensemble of running 10 models and performing majority voting. For the single-model run, the average result across 10 independent model is shown.}\label{t:ablations}
\vspace{5pt}
\scalebox{0.945}{
\begin{tabular}{|c||c|c|c|c|c|c|c|c|}
\hline
\textbf{Model and setting} & \textbf{Acc} & \textbf{F1} & \textbf{Prec} & \textbf{Rec}  & \textbf{Prec-gz} & \textbf{Prec-no} & \textbf{Rec-gz} & \textbf{Rec-no} \\ \hline \hline
\multirow{2}{*}{\textbf{Main}} & 0.797 & 0.794 & 0.810 & 0.795 & 0.750 & 0.870 & 0.900 & 0.690 \\ \cline{2-9}
\multirow{2}{*}{}              & 0.770 & 0.765 & 0.791 & 0.768 & 0.718 & 0.864 & 0.903 & 0.633 \\ \hline
\multirow{2}{*}{\textbf{Single-model}}  & 0.771 & 0.770 & 0.773 & 0.770 & 0.754 & 0.793 & 0.817 & 0.724 \\ \cline{2-9}
\multirow{2}{*}{}                            & 0.721 & 0.717 & 0.733 & 0.720 & 0.690 & 0.775 & 0.823 & 0.617 \\ \hline
\multirow{2}{*}{\textbf{Only-last}}  & 0.780 & 0.776 & 0.797 & 0.778 & 0.730 & 0.864 & 0.900 & 0.655 \\ \cline{2-9}
\multirow{2}{*}{}                            & 0.754 & 0.747 & 0.779 & 0.752 & 0.700 & 0.857 & 0.903 & 0.600 \\ \hline
\multirow{2}{*}{\textbf{No-poly}} & 0.746 & 0.738 & 0.771 & 0.743 & 0.692 & 0.850 & 0.900 & 0.586 \\ \cline{2-9}
\multirow{2}{*}{}                      & 0.672 & 0.668 & 0.678 & 0.670 & 0.649 & 0.708 & 0.774 & 0.567 \\ \hline
\multirow{2}{*}{\textbf{Poly-input}} & 0.797 & 0.794 & 0.810 & 0.795 & 0.750 & 0.870 & 0.900 & 0.690  \\ \cline{2-9}
\multirow{2}{*}{}                         & 0.672 & 0.670 & 0.675 & 0.671 & 0.657 & 0.692 & 0.742 & 0.600 \\ \hline
\multirow{2}{*}{\textbf{No-temp-aug}} & 0.797 & 0.794 & 0.810 & 0.795 & 0.750 & 0.870 & 0.900 & 0.690 \\ \cline{2-9}
\multirow{2}{*}{}                          & 0.738 & 0.732 & 0.755 & 0.735 & 0.692 & 0.818 & 0.871 & 0.600 \\ \hline
\multirow{2}{*}{\textbf{No-RGB}} & 0.797 & 0.795 & 0.802 & 0.795 & 0.765 & 0.840 & 0.867 & 0.724 \\ \cline{2-9}
\multirow{2}{*}{}                     & 0.734 & 0.729 & 0.766 & 0.735 & 0.683 & 0.850 & 0.903 & 0.567 \\ \hline
\multirow{2}{*}{\textbf{No-RGB-no-veg}} & 0.797 & 0.794 & 0.810 & 0.795 & 0.750 & 0.870 & 0.900 & 0.690 \\ \cline{2-9}
\multirow{2}{*}{}                            & 0.689 & 0.680 & 0.706 & 0.686 & 0.650 & 0.762 & 0.839 & 0.533 \\ \hline
\multirow{2}{*}{\textbf{Only-RGB+veg}} & 0.661 & 0.656 & 0.667 & 0.659 & 0.639 & 0.670 & 0.767 & 0.552 \\ \cline{2-9}
\multirow{2}{*}{}                            & 0.607 & 0.588 & 0.624 & 0.603 & 0.581 & 0.667 & 0.806 & 0.400 \\ \hline
\end{tabular}}
\end{table}

\end{document}